\documentclass[conference]{IEEEtran}
\IEEEoverridecommandlockouts
\usepackage{cite}
\usepackage{amsmath,amssymb,amsfonts}
\usepackage{algorithmic}
\usepackage{graphicx}
\usepackage{textcomp}
\usepackage{xcolor}
\def\BibTeX{{\rm B\kern-.05em{\sc i\kern-.025em b}\kern-.08em
    T\kern-.1667em\lower.7ex\hbox{E}\kern-.125emX}}

\makeatletter
\newcommand{\linebreakand}{%
\end{@IEEEauthorhalign}
\hfill\mbox{}\par
\mbox{}\hfill\begin{@IEEEauthorhalign}
}
\makeatother
\begin{document}

\title{A Cross-direction Task Decoupling Network for Small Logo Detection}

\author{
	\IEEEauthorblockN{Sujuan Hou\IEEEauthorrefmark{1}, Xingzhuo Li\IEEEauthorrefmark{1}, Weiqing Min\IEEEauthorrefmark{2}, Jiacheng Li\IEEEauthorrefmark{1}, Jing Wang\IEEEauthorrefmark{1}, Yuanjie Zheng\IEEEauthorrefmark{1}, Shuqiang Jiang\IEEEauthorrefmark{2}}
	\IEEEauthorblockA{\IEEEauthorrefmark{1} School of Information Science and Engineering, Shandong Normal University\\ Jinan, China}
	\IEEEauthorblockA{\IEEEauthorrefmark{2} Key Laboratory of Intelligent Information Processing, Institute of Computing Technology, Chinese Academy of Sciences\\ Beijing, China}
	\IEEEauthorblockA{\{hsj1985\}@126.com, \{2020317105, 2021317140\}@stu.sdnu.edu.cn,\\ \{minweiqing, sqjiang\}@ict.ac.cn, \{jingwang1551\}@163.com, \{zhengyuanjie\}@gmail.com}
	\thanks{
		This research was funded by the National Natural Science Foundation of China (62072289 and U19B2040), and CAAI-Huawei MindSpore Open Fund. 
		
		Corresponding author: Weiqing Min.}
}

\maketitle

\begin{abstract}
	Logo detection plays an integral role in many applications. However, handling small logos is still difficult since they occupy too few pixels in the image, which burdens the extraction of discriminative features. The aggregation of small logos also brings a great challenge to the classification and localization of logos. To solve these problems, we creatively propose Cross-direction Task Decoupling Network (CTDNet) for small logo detection. We first introduce Cross-direction Feature Pyramid (CFP) to realize cross-direction feature fusion by adopting horizontal transmission and vertical transmission. In addition, Multi-frequency Task Decoupling Head (MTDH) decouples the classification and localization tasks into two branches. A multi-frequency attention convolution branch is designed to achieve more accurate regression by combining discrete cosine transform and convolution creatively. Comprehensive experiments on four logo datasets demonstrate the effectiveness and efficiency of the proposed method.
\end{abstract}

\begin{IEEEkeywords}
object detection, logo detection, multi-scale feature, attention
\end{IEEEkeywords}

\section{Introduction}
Logo detection is a special application of object detection in computer vision. It has drawn increasing attention for its various practical applications such as intelligent transportation, trademark infringement detection, and multimedia information collection and analysis.

\begin{figure}[!h]
	\centering
	\includegraphics[width=0.5\textwidth]{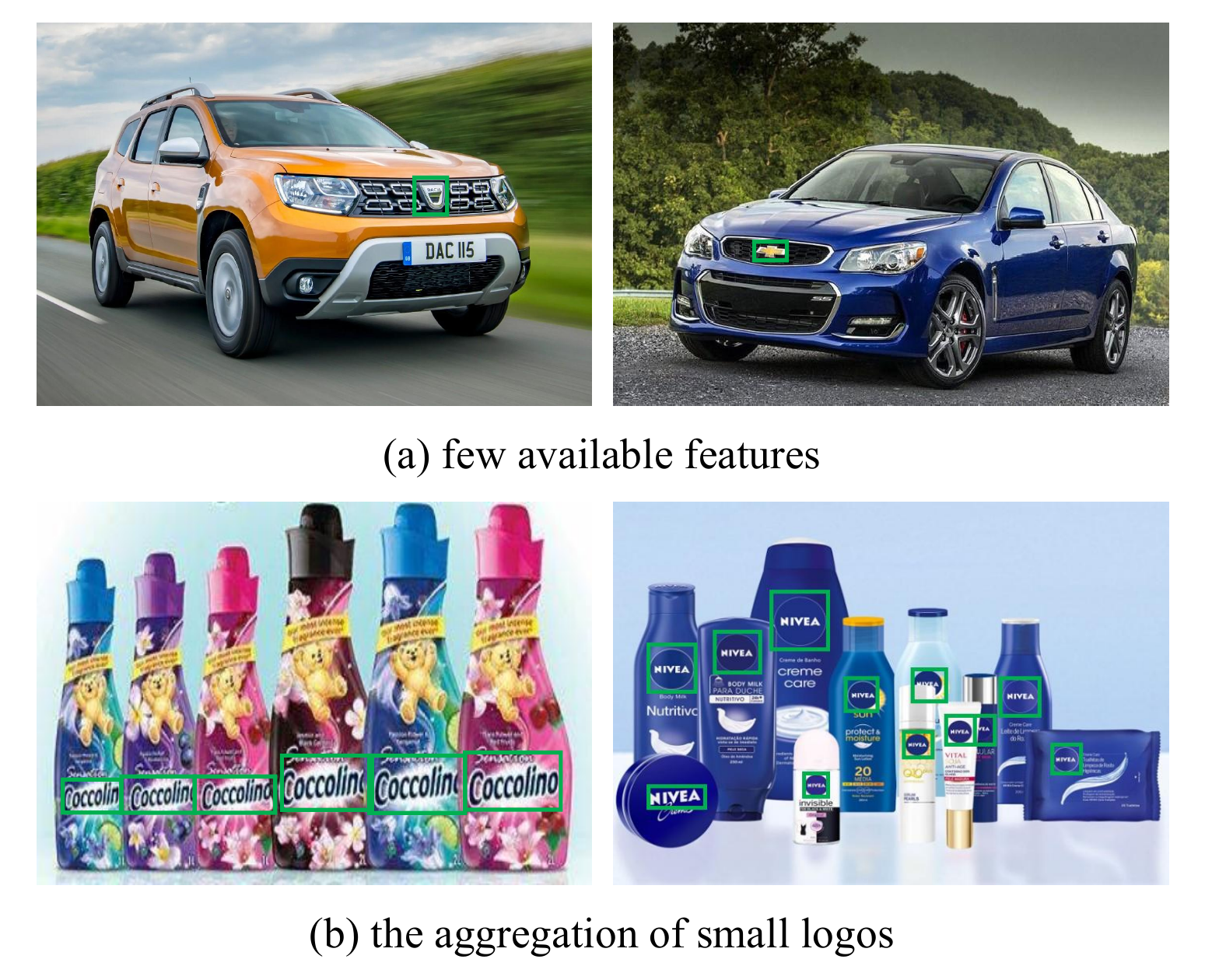}
	\caption{Small logo detection challenges: (a) small logos usually occupy too few pixels. (b) the aggregation of multiple small logos.}
	\label{fig:introduction}
\end{figure}

With the increasing number of brands, the research of small logo detection algorithms still remains extremely challenging. (1) The first difficulty mainly lies in the small object regions, which has high requirements for feature extraction. Two car logos have similar backgrounds and logo shapes, and the small logo exacerbates the detection difficulties in Fig. \ref{fig:introduction} (a). It is difficult to extract distinguishing features from small logos since they have fewer features available and are more susceptible to external interference. (2) The second challenge is that the aggregation of small logos causes the difficulty of regression and classification. Due to practical needs such as merchandising and advertising shooting, the aggregation detection of logos is a widely occurring challenge in logo detection. An example of aggregation is shown in Fig. \ref{fig:introduction} (b). The two well-liked categories ‘Coccolino’ and ‘NIVEA’ are extensively concentrated. The bounding box of the aggregated region is too close to convergence and regression in logo detection.

The existing algorithms have been studied for small object detection, among which the feature fusion method has shown high performance \cite{lin2017feature,qiao2021detectors,pang2019libra}. In addition, many deep learning researchers have explored classification and localization \cite{zhu2019feature,wang2020side,wu2020rethinking}. However, there are three main problems that need to be solved when existing models are directly used to small logo detection. Firstly, the previous work mainly focused on the object detection of natural scene images, which often failed to achieve satisfactory results when it was directly applied to the small logo. Secondly, few existing studies have specifically analyzed the difficulty of small logo detection, while we analyzed it from two perspectives of few extractable features and logo aggregation. Finally, existing models are more suited to generic object detection, making it difficult to achieve a proper trade-off between accuracy and speed when they are used in small logo detection.

In this paper, we propose a novel logo detection method Cross-direction Task Decoupling Network (CTDNet). It utilizes creative Cross-direction Feature Pyramid (CFP) to achieve more effective feature fusion in small logo regions. To solve the aggregation of small logos, the network decouples classification and localization tasks into two branches by introducing Multi-frequency Task Decoupling Head (MTDH). In CFP, we rethink the flow of feature information, adopting horizontal transmission and vertical transmission. An iterative feature pyramid is adopted for horizontal transmission, where the output features of the former pyramid are the good feature distribution of the latter pyramid. The vertical transmission after the horizontal transmission is focused on extracting balanced semantic representations, which enhances the feature information at each resolution. In MTDH, the fully-connected layer is utilized to build fully connected classification branch since it has a stronger ability to distinguish logo categories. To solve the aggregation difficulty of small logos, our proposed multi-frequency attention convolution branch complements the advantages of discrete cosine transform and convolution. Discrete cosine transform completes the focusing and extraction of important information in the image, while the feature transformation process of the convolutional layer determines its good sensitivity to position.

The main contributions of this paper can be summarized as follows:
\begin{itemize}
	\item We propose Cross-direction Feature Pyramid (CFP) to build a simple and effective pyramid through horizontal transmission and vertical transmission.
	\item A multi-frequency attention convolution branch is designed to solve the logo aggregation difficulty by combining discrete cosine transform and convolution in Multi-frequency Task Decoupling Head (MTDH).
	\item We conduct extensive experiments on four benchmark logo datasets, including FlickrLogos-32, QMUL-OpenLogo, FoodLogoDet-1500, and LogoDet-3K. The experimental results demonstrate the effectiveness of the proposed model.
\end{itemize}

\begin{figure*}[!h]
	\centering
	\includegraphics[width=1\textwidth]{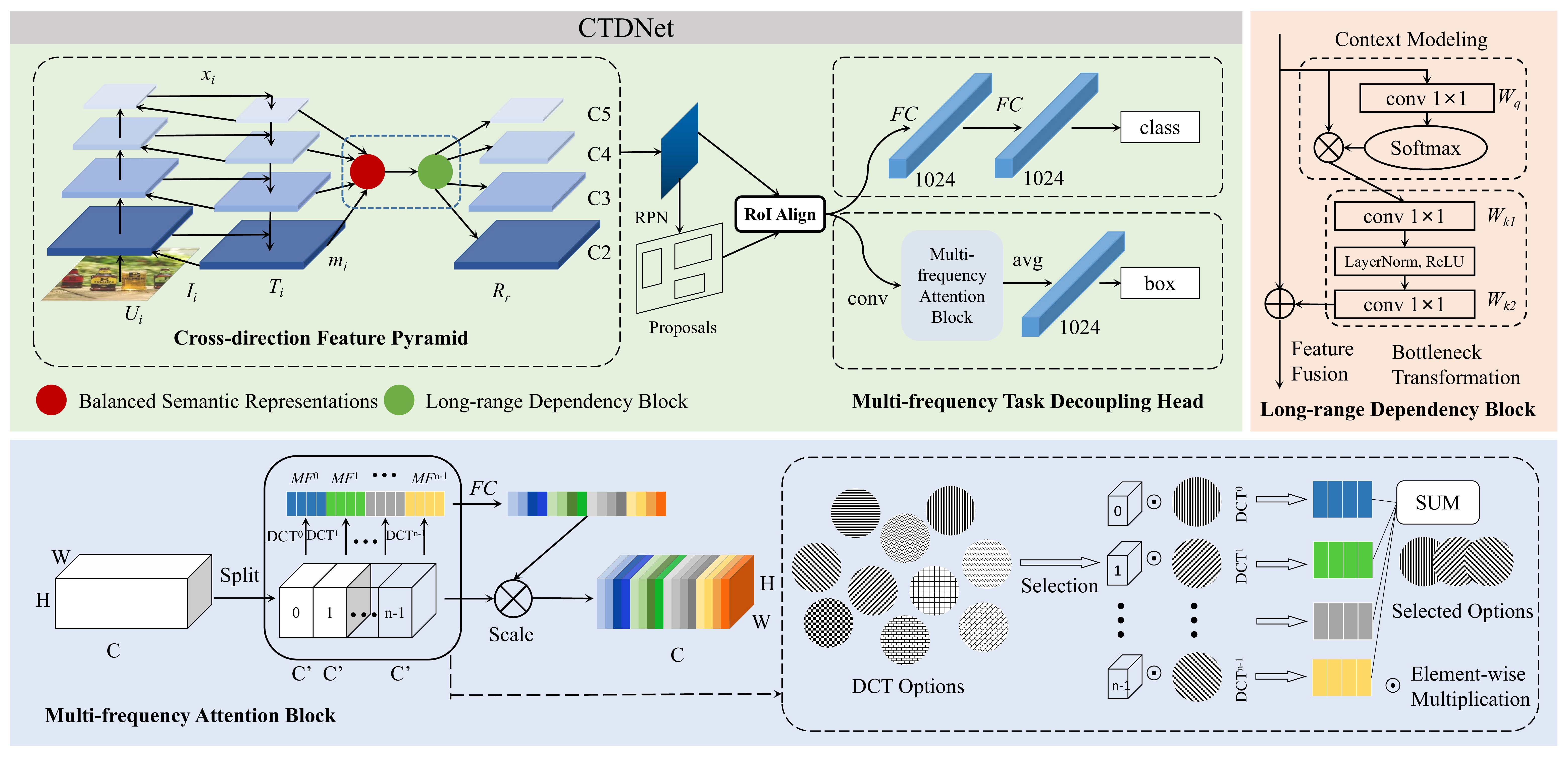}
	\caption{Overview of proposed CTDNet for small logo detection. In multi-frequency attention block, DCT denotes Discrete Cosine Transform. For simplicity, DCT indices are represented in the one-dimensional format.}
	\label{fig:model}
\end{figure*}

\section{Related Work}

Object detection is one of the most basic problems in computer vision. In the era of deep learning, object detection is divided into two genres: two-stage and one-stage. The two-stage algorithm generates regional proposals based on the image content and then performs classification and bounding box localization \cite{wu2020rethinking,zhang2020dynamic,sun2021sparse}. The one-stage algorithm is characterized by generating the category and localization coordinates directly \cite{zhang2020bridging,li2020generalized,ge2021yolox}. 

Logo detection has been extensively researched in many realistic applications. Early methods for logo detection generally relied on manual feature extraction techniques and traditional classification models. Recently, a series of deep logo detection methods as well as large-scale datasets have been proposed by exploiting the state-of-the-art object detection models \cite{wang2021cross,wang2022logodet,hou2021foodlogodet,hou2022deep}. Logo-Yolo \cite{wang2022logodet} was proposed to solve imbalanced samples of logos, and a high-quality logo dataset LogoDet-3K was built. MFDNet \cite{hou2021foodlogodet} was designed to address the multi-scale and similar logo difficulties in food logo detection, and a large dataset FoodLogoDet-1500 was constructed to solve data limitations. A cross-view learning method \cite{wang2021cross} provided ideas for logo detection. These methods promote the development of logo detection research, especially the detection problems in specific scenarios, such as the multi-scale and similarity of logo. Meanwhile, a series of high quality logo datasets have been established, which greatly facilitates future research work. However, few studies have focused on detection of small logos. Our work provides a solution and reference for small logo detection.

\section{Method}
In this section, we present a logo detection method CTDNet shown in Fig. \ref{fig:model}. After extracting the basic features from the input image, the model first inputs the feature map into CFP to learn multi-scale features. Then the feature map is fed into Region Proposal Network (RPN) to obtain region proposals. Finally, the model is sent to MTDH for classification and localization. All components will be described in detail in the following sections. 

\subsection{Cross-direction Feature Pyramid}
\subsubsection{Horizontal Transmission}
In this subsection, we first define the basic structure of FPN. $S$ is the number of stages, e.g., $S$ = 4. In Fig. \ref{fig:model}, ${x}_{i}$ is the connection of two pyramids. ${m}_{i}$ ($ \forall i = 1, ..., S$) denotes a set of output feature maps, and is defined as:
\begin{align}
	{m}_{i}={T}_{i}\left({m}_{i+1}, {x}_{i}\right), {x}_{i}={U}_{i}\left({x}_{i-1}\right)
\end{align}
where ${T}_{i}$ represents the $i\mbox{-}th$ top-down FPN operation and ${U}_{i}$ indicates the $i\mbox{-}th$ stage of the bottom-up backbone. 

Our horizontal transmission adds feedback connections to FPN to form an iterative feature pyramid. As in Fig. \ref{fig:model}, the output features of one pyramid are sent to the next one. The output feature ${m}_{i}$ of iterative feature pyramid is defined as:
\begin{align}
	{m}_{i}={T}_{i}\left({m}_{i+1}, {x}_{i}\right), {x}_{i}={U}_{i}\left({x}_{i-1}, {I}_{i}\left({m}_{i}\right)\right)
\end{align}
where ${I}_{i}$ denotes the feature of the feedback connection to the backbone. Further, we expand the iterative feature pyramid as a sequential network. $N$ is the number of expansions:
\begin{align}
	{m}_{i}^{n}={T}_{i}^{n}\left({m}_{i+1}^{n}, {x}_{i}^{n}\right), {x}_{i}^{n}={U}_{i}^{n}\left({x}_{i-1}^{n}, {I}_{i}^{n}\left({m}_{i}^{n-1}\right)\right)
\end{align}

\subsubsection{Vertical Transmission}
After the horizontal transmission, we extract the balanced semantic representation on the vertical transmission to strengthen the relationship between levels. In Fig. \ref{fig:model}, we first resize all the feature maps to a medium size to obtain balanced features $R$:
\begin{align}
	R=\frac{1}{S} \sum_{r=r_{min}}^{r_{max}} R_{r}
\end{align}
where $S$ is the number of feature levels. The bottom and top resolution levels are labeled as $r_{min}$ and $r_{max}$, respectively. $R_r$ denotes the feature level with a resolution of $r$. Afterward, the obtained features are rescaled using the same but opposite way to complete the reinforcement.

To further optimize the extracted features, we capture the long-range dependency of the balanced features using global context modeling \cite{cao2019gcnet} in Fig. \ref{fig:model}.

\subsection{Multi-frequency Task Decoupling Head}
After RoI Align, the localization and classification tasks are performed respectively by two different branches. In the localization branch, the feature transformation process is implemented using convolution to extract more discriminative features. Besides, the channel representation can be considered as a compression process using frequency analysis \cite{qin2021fcanet}. The channel informations are compactly encoded to maintain their representational capability. We use the discrete cosine transform to compress channels and take its multiple frequency components as the channel attention mechanism.

The multi-frequency attention block is shown in Fig. \ref{fig:model}. We denote ${MF}^{i}$ for a particular frequency component, which can be considered as channel attention pre-processing of the feature map $X$:
\begin{align}
	{MF}^{i}=\sum_{h=0}^{H-1} \sum_{w=0}^{W-1} X^{i} B_{h, w}^{i}
\end{align}
where $i$ denotes the $i\mbox{-}th$ group. $(h,w)$ is the position of the 2D image. $B$ is the basic function of the discrete cosine transform.

The complete vector $MF$ can be obtained by the concatenation of multiple frequency components:
\begin{align}
	{MF}={concat}\left(\left[{MF}^{0}, {MF}^{1}, \cdots, {MF}^{n-1}\right]\right)
\end{align}

The final multi-frequency attention block can be written as:
\begin{align}
	{MFB}={sigmoid}({FC}({MF}))
\end{align} 

In the classification task, fully connected classification branch is adopted to strengthen the sensitivity of the logo category by combining two fully connected layers. 

\subsection{Loss Function}
In the CTDNet, the final loss function consists of ${L}_{rpn}$, ${L}_{fc}$ and ${L}_{conv}$:
\begin{align}
	{L}={L}_{rpn}+\omega_{fc}{L}_{fc}+\omega_{conv}{L}_{conv}
\end{align}
where ${L}_{rpn}$, ${L}_{fc}$ and ${L}_{conv}$ are the losses for RPN, fully connected classification branch and multi-frequency attention convolution branch, respectively. $\omega_{fc}$ and $\omega_{conv}$ are the weights of the classification branch and the localization branch, respectively. 

We realize ${L}_{conv}$ by Smooth $L1$ Loss and ${L}_{fc}$ by Cross-Entropy Loss function. 

\section{Experiments}

\begin{table*}[!h]
	\caption{Statistics of four logo datasets.}
	\centering
	\label{tab:datasets}
	\scalebox{1}
	{
		\begin{tabular}{ccccccc}
			\hline
			\textbf{Datasets} & \textbf{\#Classes} & \textbf{\#Images} & \textbf{\#Objects} & \textbf{\#Trainval} & \textbf{\#Test} & \textbf{\#Small Objects} \\
			\hline
			FlickrLogos-32 \cite{romberg2011scalable} & 32 & 2,240 & 3,405 & 1,478 & 762 & 185 \\
			QMUL-OpenLogo \cite{su2018open} & 352 & 27,083 & 51,207 & 18,752 & 8,331 & 11,841 \\
			FoodLogoDet-1500 \cite{hou2021foodlogodet} & 1,500 & 99,768 & 145,400 & 80,280 & 19,488 & 16,463 \\
			LogoDet-3K \cite{wang2022logodet} & 3,000 & 158,652 & 194,261 & 142,142 & 16,510 & 3,508\\
			\hline
		\end{tabular}
	}
\end{table*}

\begin{table*}[!h]
	\caption{Detection results on four datasets (\%).}
	\centering
	\label{tab:results}
	\scalebox{0.8}
	{
		\begin{tabular}{cccccccccccccccccc}
			\hline
			Method & Backbone & mAP & $AP_S$ & $AP_M$ & $AP_L$ & mAP & $AP_S$ & $AP_M$ & $AP_L$ & mAP & $AP_S$ & $AP_M$ & $AP_L$ & mAP & $AP_S$ & $AP_M$ & $AP_L$ \\
			\hline
			&  & \multicolumn{4}{c}{FlickrLogos-32} & \multicolumn{4}{c}{QMUL-OpenLogo} & \multicolumn{4}{c}{FoodLogoDet-1500} & \multicolumn{4}{c}{LogoDet-3K} \\
			\cline{3-18}
			One-stage: \\
			FSAF \cite{zhu2019feature} & ResNet-50-FPN & 84.9 & 26.5 & 78.9 & 89.9 & 48.7 & 31.0 & 50.4 & 60.0 & 83.4 & 69.4 & 71.8 & 78.5 & 63.0 & 26.4 & 56.0 & 67.1 \\
			ATSS \cite{zhang2020bridging} & ResNet-50-FPN & 84.0 & 21.3 & 73.8 & 89.7 & 47.3 & 29.6 & 49.5 & 58.8 & 84.2 & 66.5 & 72.0 & 77.8 & 68.3 & 20.3 & 56.4 & 73.9 \\
			GFL \cite{li2020generalized} & ResNet-50-FPN & 84.4 & 20.5 & 80.0 & 89.8 & 47.5 & 30.4 & 48.5 & 58.9 & 74.3 & 64.6 & 72.6 & 77.7 & 60.1 & 14.7 & 50.8 & 65.1 \\
			TOOD \cite{feng2021tood} & ResNet-50-FPN & 87.7 & 19.3 & 84.5 & 93.9 & 51.9 & 33.3 & 52.7 & 63.6 & 82.7 & 74.0 & 79.9 & 86.5 & 78.5 & 40.0 & 73.3 & 81.9 \\ 
			DW \cite{li2022dual} & ResNet-50-FPN & 88.6 & 24.7 & 83.0 & 94.4 & 54.9 & 35.3 & 55.2 & 69.0 & 82.2 & 76.2 & 78.4 & 86.3 & 85.6 & 54.5 & 79.4 & 88.9 \\
			\hline
			Two-stage: \\
			Faster R-CNN \cite{ren2015faster} & ResNet-50-FPN & 87.3 & 16.8 & 82.4 & 94.5 & 53.3 & 33.7 & 53.7 & 67.2 & 84.0 & 76.0 & 81.0 & 88.4 & 85.2 & 51.5 & 79.7 & 88.3 \\
			Cascade R-CNN \cite{cai2018cascade} & ResNet-50-FPN & 87.7 & 14.3 & 83.3 & 94.6 & 52.4 & 30.9 & 51.6 & 65.4 & 83.3 & 74.3 & 79.0 & 87.5 & 84.5 & 43.7 & 77.1 & 88.0 \\
			PANet \cite{liu2018path} & ResNet-50-PAFPN & 87.5 & 17.1 & 82.7 & 93.8 & 53.7 & 29.3 & 53.9 & 68.3 & 84.0 & 75.7 & 81.6 & 88.1 & 85.2 & 48.4 & 78.9 & 88.5 \\
			Libra R-CNN \cite{pang2019libra} & ResNet-50-BFP & 88.6 & 22.1 & 87.0 & 94.2 & 56.5 & 35.1 & 57.5 & 69.3 & 83.3 & 76.5 & 80.6 & 87.2 & 82.6 & 52.2 & 77.7 & 85.8 \\
			Generalized IoU \cite{rezatofighi2019generalized} & ResNet-50-FPN & 86.5 & 18.9 & 82.1 & 93.6 & 52.4 & 31.3 & 52.9 & 65.5 & 83.5 & 74.3 & 80.6 & 87.7 & 84.2 & 48.4 & 78.9 & 87.4 \\
			Complete IoU \cite{zheng2020distance} & ResNet-50-FPN & 88.2 & 17.7 & 86.3 & 94.2 & 52.7 & 31.2 & 53.0 & 66.4 & 83.3 & 75.0 & 80.9 & 87.1 & 83.9 & 51.0 & 78.6 & 86.9 \\
			Dynamic R-CNN \cite{zhang2020dynamic} & ResNet-50-FPN & 87.6 & 23.2 & 82.7 & 94.5 & 53.1 & 30.9 & 53.5 & 66.5 & 84.4 & 75.7 & 81.2 & 88.6 & 87.7 & 56.2 & 82.1 & 90.6 \\ 
			Double-head R-CNN \cite{wu2020rethinking} & ResNet-50-FPN & 88.2 & 23.1 & 86.4 & 94.3 & 54.2 & 32.9 & 54.3 & 67.3 & 85.5 & 78.3 & 82.2 & \textbf{89.6} & 86.4 & 48.3 & 79.8 & 89.8 \\
			SABL \cite{wang2020side} & ResNet-50-FPN & 87.6 & 10.1 & 86.9 & \textbf{94.9} & 54.8 & 33.1 & 55.3 & 68.8 & 83.3 & 72.4 & 79.4 & 88.2 & 85.1 & 42.2 & 78.7 & 88.7 \\
			Sparse R-CNN \cite{sun2021sparse} & ResNet-50-FPN & 80.1 & 20.2 & 71.0 & 87.5 & 52.2 & 35.3 & 53.9 & 63.9 & 81.8 & 77.5 & 79.9 & 85.7 & 37.9 & 40.3 & 46.4 & 40.7 \\
			Guided Anchoring \cite{wang2019region} & ResNet-50-FPN & 86.8 & 11.7 & 82.5 & 93.5 & 52.2 & 32.7 & 53.7 & 65.9 & 85.4 & 77.2 & 82.6 & 89.0 & 86.3 & 56.5 & 81.6 & 89.4 \\
			\textbf{CTDNet(ours)} & \textbf{ResNet-50-CFP} & \textbf{89.7} & \textbf{34.3} & \textbf{89.0} & 94.3 & \textbf{58.4} & \textbf{37.6} & \textbf{60.7} & \textbf{71.7} & \textbf{85.6} & \textbf{79.8} & \textbf{83.1} & 88.8 & \textbf{88.2} & \textbf{58.4} & \textbf{82.9} & \textbf{90.8} \\
			\hline
		\end{tabular}
	}
\end{table*}

\begin{table*}[!h]
	\caption{Evaluating individual component on four datasets (\%).}
	\centering
	\label{tab:ablation}
	\begin{tabular}{cccccccccc}
		\hline
		Method & Backbone & mAP & $AP_S$ & $AP_M$ & $AP_L$ & mAP & $AP_S$ & $AP_M$ & $AP_L$\\
		\hline
		&  & \multicolumn{4}{c}{FlickrLogos-32} & \multicolumn{4}{c}{QMUL-OpenLogo} \\
		\cline{3-10}
		Faster R-CNN & ResNet-50-FPN & 87.3 & 16.8 & 82.4 & 94.5 & 53.3 & 33.7 & 53.7 & 67.2 \\
		Faster R-CNN+CFP & ResNet-50-CFP & 88.5 & 24.2 & 87.6 & 94.6 & 57.3 & 36.1 & 58.5 & 70.1 \\
		Faster R-CNN+MTDH & ResNet-50-FPN & 88.6 & 21.6 & 86.5 & \textbf{94.7} & 55.5 & 34.8 & 56.2 & 67.6 \\
		Faster R-CNN+CFP+MTDH & ResNet-50-CFP & \textbf{89.7} & \textbf{34.3} & \textbf{89.0} & 94.3 & \textbf{58.4} & \textbf{37.6} & \textbf{60.7} & \textbf{71.7} \\
		\hline
		&  & \multicolumn{4}{c}{FoodLogoDet-1500} & \multicolumn{4}{c}{LogoDet-3K} \\
		\cline{3-10}
		Faster R-CNN & ResNet-50-FPN & 84.0 & 76.0 & 81.0 & 88.4 & 85.2 & 51.5 & 79.7 & 88.3 \\
		Faster R-CNN+CFP & ResNet-50-CFP & 85.5 & 77.1 & 82.5 & \textbf{89.6} & 85.9 & 55.9 & 81.1 & 88.9 \\
		Faster R-CNN+MTDH & ResNet-50-FPN & 85.1 & 77.6 & 82.1 & 88.9 & 87.2 & \textbf{58.5} & 81.8 & 90.1 \\
		Faster R-CNN+CFP+MTDH & ResNet-50-CFP & \textbf{85.6} & \textbf{79.8} & \textbf{83.1} & 88.8 & \textbf{88.2} & 58.4 & \textbf{82.9} & \textbf{90.8} \\
		\hline
	\end{tabular}
\end{table*}

\begin{table}[!h]
	\caption{Logo method detection results on two datasets (\%).}
	\centering
	\label{tab:logomethods}
	\scalebox{0.9}
	{
		\begin{tabular}{cccc}
			\hline
			Method & Backbone & FlickrLogos-32 & QMUL-OpenLogo \\
			\hline
			Logo-Yolo \cite{wang2022logodet} & DarkNet-53 & 76.1& 53.2 \\
			OSF-Logo \cite{meng2021adaptive} & ResNet-50-FPN & 87.0 & 53.3 \\
			MFDNet \cite{hou2021foodlogodet} & ResNet-50-BFP & 86.2 & 51.3\\
			DSFP-GA \cite{zhang2023discriminative} & ResNet-50-DSFP & 87.1 & 54.0 \\
			\textbf{CTDNet(ours)} & \textbf{ResNet-50-CFP} & \textbf{89.7} & \textbf{58.4} \\
			\hline
		\end{tabular}
	}
\end{table}

\begin{table}[!h]
	\caption{Experiments on parameter sensitivity.}
	\centering
	\label{tab:sensitivity}
	\scalebox{1}
	{
		\begin{tabular}{cccc}
			\hline
			Method & Group Number & Loss Weight & mAP(\%) \\
			\hline
			Faster R-CNN(baseline) & - & - & 87.3 \\
			Faster R-CNN+MTDH & c4f2 & 0.5 & 87.9 \\
			Faster R-CNN+MTDH & c4f2 & 1.0 & 87.8 \\
			Faster R-CNN+MTDH & c4f2 & 2.0 & 88.6 \\
			Faster R-CNN+MTDH & c6f4 & 2.0 & 88.4 \\
			CTDNet(ours) & c4f2 & 0.5 & 86.7 \\
			CTDNet(ours) & c4f2 & 1.0 & 88.6 \\
			CTDNet(ours) & c4f2 & 2.0 & 89.7 \\
			CTDNet(ours) & c6f4 & 2.0 & 89.2 \\
			\hline
		\end{tabular}
	}
\end{table}

\subsection{Experimental Setting}
To evaluate the effectiveness of the proposed CTDNet, we complete comprehensive experimental validation on four datasets. They include two small-scale datasets FlickrLogos-32 \cite{romberg2011scalable} and QMUL-OpenLogo \cite{su2018open}, the medium-scale food dataset FoodLogoDet-1500 \cite{hou2021foodlogodet}, and the large-scale dataset LogoDet-3K \cite{wang2022logodet}. The detailed description of these datasets is shown in Table \ref{tab:datasets}.

We implement our method based on the publicly available MMDetection toolbox \cite{chen2019mmdetection}. For evaluation, we use the widely used mean Average Precision (mAP) \cite{everingham2010pascal}, with an IoU threshold of 0.5. Considering different sizes of logos, we adopt $AP_S$, $AP_M$, $AP_L$ respectively, where $AP_S$ is the Average Precision (AP) for small logo objects (area $<$ $32^2$), $AP_M$ is the AP for medium logo objects ($32^2$ $<$ area $<$ $96^2$), and $AP_L$ is the AP for large logo objects (area $>$ $96^2$). In our experiments, the basic detection network is trained using Stochastic Gradient Descent (SGD), and the initial learning rate is set to 0.002. The input images are resized to 1000 × 600, the weight decay is 0.0001, and the momentum is 0.9. We follow the settings in MMDetection for other hyperparameters.

\begin{figure*}[!h]
	\centering
	\includegraphics[width=1\textwidth]{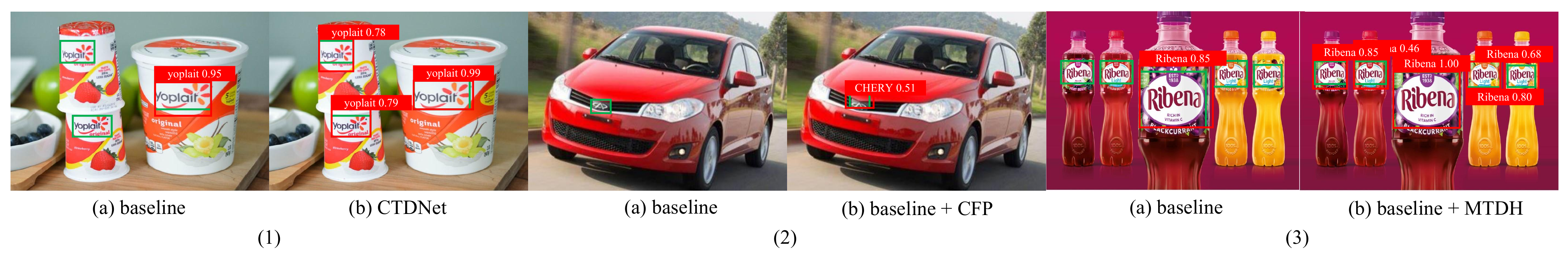}
	\caption{Comparison of visualization results between baseline and CTDNet. Green box: ground-truth box. Red box: the location of the detected logo.}
	\label{fig:vis}
\end{figure*}

\begin{figure*}[!h]
	\centering
	\includegraphics[width=1\textwidth]{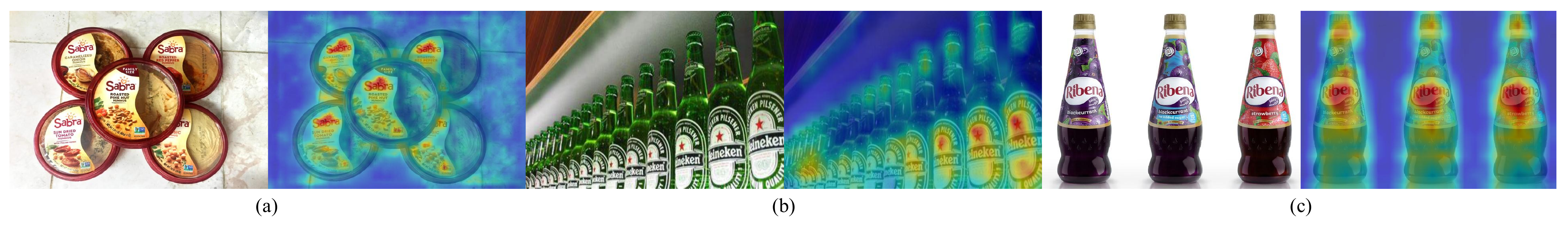}
	\caption{Heatmap visualization by cross-direction feature pyramid.}
	\label{fig:heatmap}
\end{figure*}

\begin{figure}[!h]
	\centering
	\includegraphics[width=0.5\textwidth]{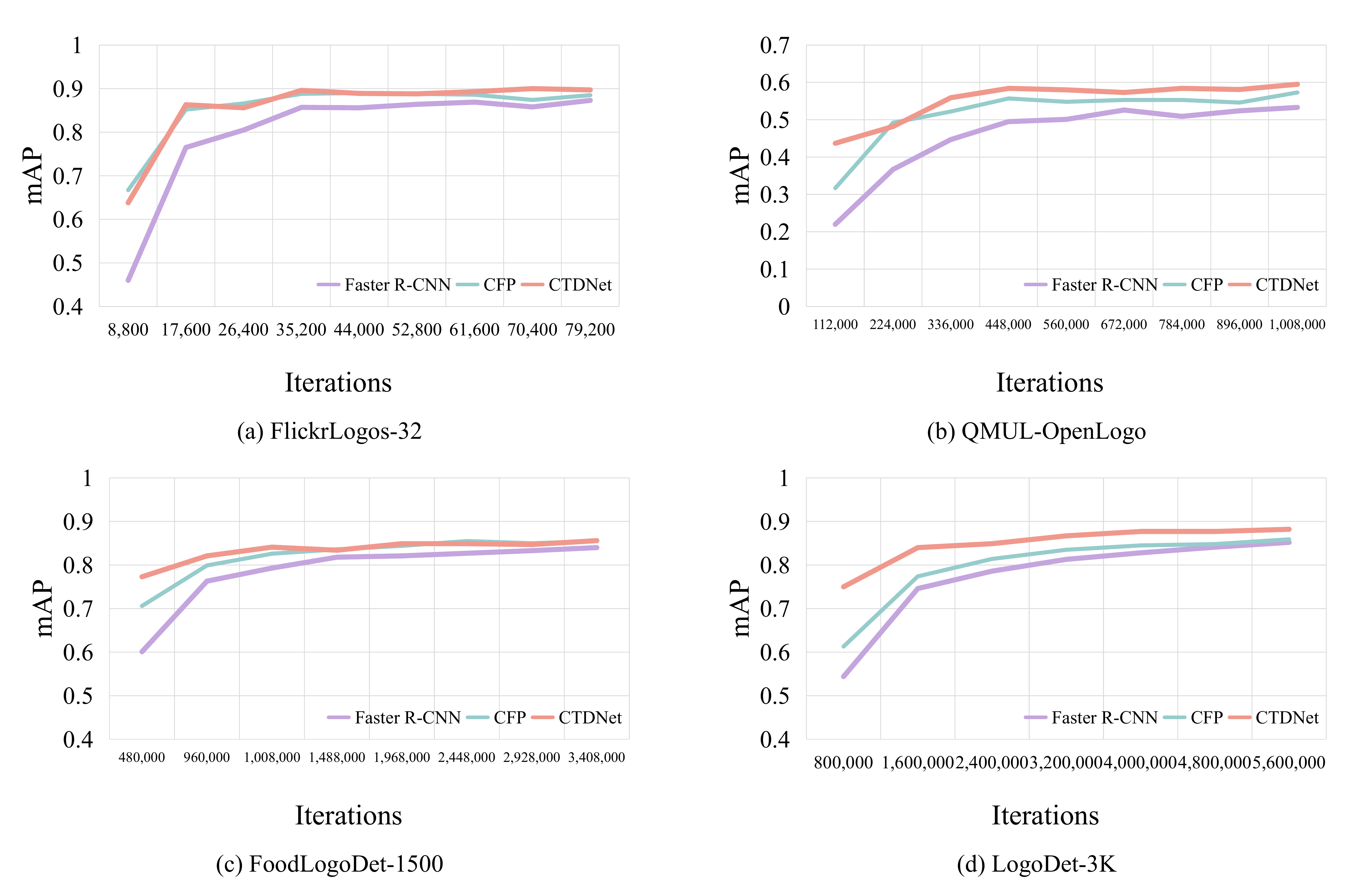}
	\caption{Comparison between different strategies with increasing number of iterations.}
	\label{fig:iterations}
\end{figure}

\subsection{Main Results}
In this subsection, we show the main results of CTDNet conducted on four datasets. To validate the generality of the proposed CTDNet, we compare the proposed model with several other popular baselines, including one-stage series and two-stage series, as reported in Table \ref{tab:results}. For a fair comparison with other detectors, we equip all baselines with ResNet-50 and FPN as the backbone. On all four datasets, we observe that the proposed CTDNet method is superior to other baselines, which achieves the best performance with mAP and $AP_S$. The proposed CTDNet strategy produces dominant performance compared to other approaches utilizing feature fusion, like PANet and Libra R-CNN. To examine the effectiveness for detection head, we use Double-head R-CNN and SABL as comparison methods. Under evaluation metrics of mAP and $AP_S$, we can see that CTDNet obtains better performance than them.

In Table \ref{tab:logomethods}, more contrastive experiments with logo-detection-oriented methods are conducted on FlickrLogos-32 and QMUL-OpenLogo, including Logo-Yolo \cite{wang2022logodet}, OSF-Logo \cite{meng2021adaptive}, MFDNet \cite{hou2021foodlogodet} and DSFP-GA \cite{zhang2023discriminative}.

\subsection{Ablation Studies}
In this subsection, we show the ablation studies of CFP and MTDH on four datasets in Table \ref{tab:ablation}. The mAP, $AP_S$, $AP_M$ and $AP_L$ for accuracy are applied to the reported results. We use Faster R-CNN equipped with ResNet-50 and FPN as the baseline.

\subsubsection{Quantitative Analysis}
From Table \ref{tab:ablation}, we can see that both the CFP and MTDH gain improvement on four datasets. On LogoDet-3K, CFP gets 0.7\% mAP more than Faster R-CNN, while MTDH improves 2.0\% mAP. The combination of CFP and MTDH brings the best performance to the model. In addition, CTDNet gains 2.4\%, 5.1\%, and 1.6\% mAP improvement in comparison with baseline on FlickrLogos-32, QMUL-OpenLogo, and FoodLogoDet-1500, respectively. Both CFP and MTDH can improve $AP_S$ on four datasets, which illustrates the improvement of small logo detection performance by these two components. In comparison with the baseline, our model improves the $AP_S$ by 17.5\%, 3.9\%, 3.8\% and 6.9\% on FlickrLogos-32, QMUL-OpenLogo, FoodLogoDet-1500, and LogoDet-3K, respectively. The ablation study shows that the formulations of CFP and MTDH have the best configuration.

\subsubsection{Visualization Analysis}
Fig. \ref{fig:vis} provides the visualization results by Faster-RCNN (hereinafter referred to as baseline), ‘baseline + CFP’, ‘baseline + MTDH’ and CTDNet from the localization and accuracy of testing on LogoDet-3K. As seen in Fig. \ref{fig:vis} (1), CTDNet is far superior to baseline in both localization and accuracy. It is worth mentioning that logo categories ‘yoplaint’ and ‘CHERY’ can not be detected by baseline, while our model obtains good detection results in Fig. \ref{fig:vis} (1) (2). These results demonstrate that CFP is consistent with our intuition concerning a non-negligible improvement in feature fusion. From the comparison between baseline and ‘baseline + MTDH’, as shown in Fig. \ref{fig:vis} (3), baseline only detects the largest logo in the category ‘Ribena’, while ‘baseline + MTDH’ detects all five logos. These results indicate that our detection head plays a very critical role in detecting the aggregation of small logos.

Fig. \ref{fig:heatmap} gives two illustrative examples of the heatmap visualization results by CFP. The bright colors in the figure represent that CFP extracts more representative feature information at this location. Comparing with the original image, it can be seen that the concerned positions of CFP are well aligned with detection logos. These results further prove the effectiveness of our designed CFP.

\subsubsection{Parameter Sensitivity}
Different iteration times are set to compare baseline, CFP and CTDNet in terms of convergence and accuracy. As can be seen from Fig. \ref{fig:iterations} that with the increase of iteration times, the three models all achieve performance convergence on the four data sets, among which CTDNet achieves the best performance. 

For MTDH, the comparative experiments on group numbers and loss weights are conducted. Table \ref{tab:sensitivity} shows the performance of baseline, MTDH and CTDNet with different parameter settings on FlickrLogos-32. For the loss weights, we set three values of 0.5, 1.0 and 2.0. For the number of groups, we set two types, namely c6f4 and c4f2. The former represents 6 convolutional layers and 4 fully connected layers, while the latter represents 4 convolutional layers and 2 fully connected layers. According to Table \ref{tab:sensitivity}, it can be seen that all three values of loss weights enable MTDH and CTDNet to obtain performance improvements compared to the baseline. Both MTDH and CTDNet obtain the best performance when the loss weight is 2.0 and the group number is c4f2. 

\section{Conclusion}
In this paper, we propose a logo detection model Cross-direction Task Decoupling Network (CTDNet), which introduces CFP and MTDH to address the detection difficulties caused by small logos. The proposed CFP utilizes a simple pyramid to accomplish feature fusion through horizontal transmission and vertical transmission. Meanwhile, MTDH is proposed to build different head structures for classification and localization tasks, which can solve the logo aggregation difficulty by combining discrete cosine transform and convolution. In future, we will focus on other challenges of logo detection, such as low resolution, logos with large aspect ratios and similar logos. 

\bibliographystyle{IEEEtran}
\bibliography{icme2023template}

\end{document}